# Convolutional Gated MLP: Combining Convolutions & gMLP

A. Rajagopal[+] , V. Nirmala[++]

**Abstract.** To the best of our knowledge, this is the first paper to introduce Convolutions to Gated Multi-Layer Perceptron (gMLP) and contributes an implementation of this novel Deep Learning architecture. Google Brain introduced the gMLP in May 2021. Microsoft introduced Convolutions in Vision Transformer (CvT) in Mar 2021. Inspired by both gMLP and CvT, we introduce convolutional layers in gMLP. CvT combined the power of Convolutions and Attention. Our implementation combines the best of Convolutional learning along with spatial gated MLP. Further, the paper visualizes how CgMLP learns. Visualizations show how CgMLP learns from features such as outline of a car. While Attention was the basis of much of recent progress in Deep Learning, gMLP proposed an approach that doesn't use Attention computation. In Transformer based approaches, a whole lot of Attention matrixes need to be learnt using vast amount of training data. In gMLP, the fine tunning for new tasks can be challenging by transfer learning with smaller datasets. We implement CgMLP and compares it with gMLP on CIFAR dataset. Experimental results explore the power of generalization of CgMLP, while gMLP tend to drastically overfit the training data.

To summarize, the paper contributes a novel Deep Learning architecture and demonstrates the learning mechanism of CgMLP through visualizations, for the first time in literature.

## 1 Introduction

### 1.1 Related Literature & research directions

*Potential of Transformers for multi-modal transformation tasks:*

The real power of Transformers is in its ability to model multi-modal content and its ability to transform representation from one modality to another. For example, the ability to translate an image to caption by multi-modal embedding by OSCAR is a classic example of the power of transformers. This is fundamentally because of the way a Transformer is modeled as a function of 3 trainable matrixes and the input features that maps to target features. And such transformation allow representation of visual features and language features into a common embedding, thus allowing operations in a multi-modal embedding space. With such giant encoder-decoder learning machines, transformation of multi-modal feature input to another multi-modal target output is possible.

And the availability of vast corpus of multi-modal content on the web allows for training such future multi-modal Transformers in the say way BERT was trained on the text corpus of wiki and book. OSCAR is one giant leap in architecture design towards a future of multimodal transformation tasks.

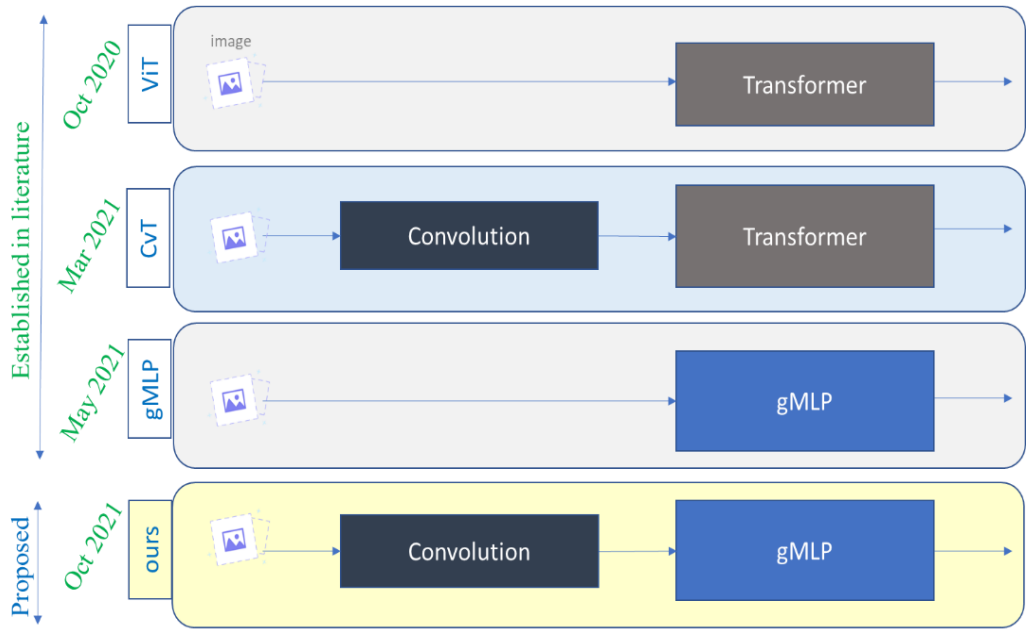

*Figure 1: Novelty*

*Journey towards Multimodal Transformers*

The steps towards a multi-modal Pre-trained Transformer model for multimodal translation is already been established. With the success of Transformer in NLP, Transformer architecture is employed in Computer Vision tasks and Visual Language tasks. For example, multiple layers of Transfomers are used in state-of-art Visual Language models such as Microsoft's OSCAR and Google's VimSIM. Another example, is Vision Transformer (ViT) for image processing tasks where an image is passed as a sequence of image patches with position embedding to Transformers, in the same way a sentence is passed as a sequence of words to Language Transformers.

*The potential of Transfer Learning in Language & Vision Transformer models:*

There is amazing progress in Deep Learning research. Since the introduction of Attention mechanism in Deep Learning architecture, Attention based modelling has been introduced in many architectures. With the advent of Transformers, the IMAGENET moment for NLP was arrived in 2018. Transformer networks is architected with layers of multi-headed attention and Feed forward networks, and often requires quite huge number of weights to be learnt. Hence training a Transformer often require significant data. But once it is trained, it can be easily adapted for new tasks. For example, Pre-trained language models such as BERT & GPT-2 enabled rapid adaptation for new custom tasks by fine tuning on small datasets with Transfer learning. Further, few shot learning was possible with 100 billion parameter Transformer networks such as GPT-3. Similar pre-training and Transfer Learning strategies are also employed by CNN networks. Both BERT and Inception are pre-trained on a large amounts of data, but can be quickly fine tunned for newer tasks with smaller datasets. The impact of Transfer Learning in both Computer Vision and Natural Language Processing has enabled widespread adoption of the AI into the mainstream.

*Strategies for learning Visual information*

Computer Vision can be modelling by learning the following
1. Long range interactions between objects in the image/video :
   By modelling the interactions across objects situation across frames of a video (or across the different regions of the image) can be realized by learning the Attention or alignment between these objects. So naturally a Transformers is apt to model such long range dependencies. So to model a video using Transformers, interactions between objects detected by Faster R-CNN across video frames can be modelling using Attention and positional embedded objects.
2. Local neighbor modelling
   Since a set of pixels in a location are often related as they represent the same object, modelling them is essential. Here, Convolutions with right kernel sizes where the receptive field is aligned to the size of the salient object will be optimal modelling strategy.
3. Common features in the image
   Features such as textures that are common across the image can be learnt by sharing the learnable parameters across the different parts of the image. This is again where Convolution layers come in handy.

*Designing better computer vision models by blending Convolutions & Attention to model both local & long-range interactions:*

Self Attention based models have low inductive biases. CNN has inductive bias in two forms. The two areas are locality and weight sharing as explained above. So Transformers based modeling is good when there is significantly large volume of data for training. Researchers introduced Convolutions in Vision Transformer to define CvT in Mar 2021. Also Facebook researchers combined Convolutional neural networks with Vision Transformers to create ConViT. By introducing a gating parameter, ConViT automatically learns to depend upon either self-attention or convolution.

*Gated MLP : Do you need attention?*

Researchers at Google Brain investigated the extend of attention that is required by experimenting with a simpler architecture in the paper "Pay Attention to MLPs". This paper tends to indicate that gMLP perform achieve same accuracy as Transformers in tasks like sentiment classification, though slightly weaker on tasks where two sentences due to lack of long range interactions.

*Comparing CNN vs Attention vs gMLP: When to use each one of them?*

|  | **CNN** | **Transformers** | **gMLP** |
|---|---|---|---|
| *Inductive biases?* | High | Minimal | Very minimal |
| *Mechanism of modelling* | Locality & Weight sharing | Long range interactions | Spatial interactions |
| *Tends to overfit on training data?* | No | Yes | Yes |
| *Transfer Learning to small datasets* | Very good | Excellent (Few shot learning) | Not established |
| *Math model* | Feature map = $X*W_{kernel}$<br>X is input image, W is CNN filter weights | Attention = $f(W_Q, W_K, W_V, X)$<br>W is dynamically generated from X. | Linear gating output = $X* (W_p X+b)$<br>W is spatial projection matrix is independent of X |
| *Good for tasks* | Classification, Image to Image transformation (e.g UNet) | Multimodal content transformation (eg. GPT2 text generation, OSCAR captioning) | Classification |

*Table 1*

Given each architecture approach has its own inherent strengths, it will be good to explore how to best leverage them.

1. *Mechanism of modelling:* While CNN is good in learning common features that can be found across video clip using weight sharing kernel, Transformers are good in learning long range interactions between objects found across different parts of the video, gMLP is good in spatial interactions.
2. *Best suited tasks:* Given the ability to model long range interactions, ideal use of the power of Transformers in translation of content from one form to another representation. Interesting the content can be multimodal in format as multimodal embedding can be represented by Transformers into another multimodal representation. So Transformers is inherently the best choice for tasks requiring multimodal transformation like Visual Question Answering. An practical application example of multimodal Transformers would be to take a grade 10 text book as a input, and let the AI simplify the concepts to be understandable by grade 6 students. Given the CNN are good at modelling spatial corelated data, it is a natural choice of any data that inherently contains spatial correlations. This could be found in images, and short video clips. For example, a spatial coreleation can be found across two images of a scene shot from two different angles such as stereoscopic cameras or pair of security cameras looking at the scene from different viewpoints. In this case of scenes from two cameras, an ideal choice would be gMLP as it dynamically models spatial interactions.

3. *Math model:* Among the 3 models under considerations, the highest amount of learnable parameters is Transformers. As shown in equations in Table 1, there are 3 large matrixes to be learnt in Transformers. Incontast, the weight sharing nature of CNN filters along it to learn faster than Transformers on relatively smaller datasets. A large number of parameters to learn also means the amount of data required to train is quite large for Transformers. This also means the tendency to over-fit on training data is high for Transformers and gMLP. The power to generalize is the aim for AI, and hence validation accuracy or accuracy after deployment is a crucial consideration. Here the over-fit nature of Transformers and gMLP means that Transformers may be ideal choice for generation tasks such as GPT2 text generation. In Transformers, the 3 Weight matrixes depend upon the input data X. In gMLP, the Weight matrix is not dependent upon input data X. This could mean that gMLP could have the potential to handle previously unseen data during inference with slightly different distribution.

*Learning to choose architecture dynamically*

The ConViT approach dynamically learns to chooses one of the two blocks (Convolutions vs Transformers) dynamically as shown in Fig 2. Actually it dynamically learns to use the best of both approaches (Convolution & Attention).

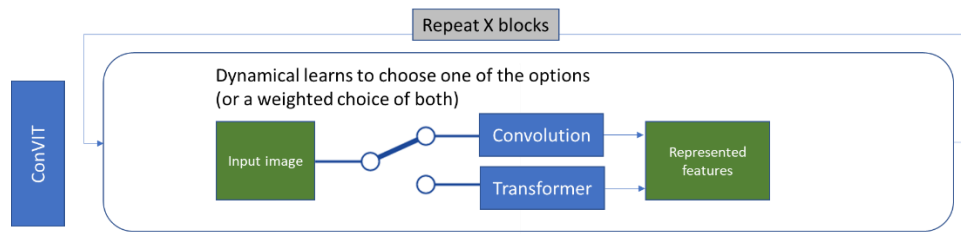

*Figure 2*

## 1.2  Research gap / Novelty

1. Oct 2020:  Computer Vision tasks is modelled with Transformer (ViT)
2. Mar 2021:  Convolution is introduced to ViT (CViT)

   3. Mar 2021: Dynamic blend of Convolution & Transformer (ConViT)
   4. May 2021: gMLP is proposed as performs comparably to Transformer
   5. This work: Introduces Convolution to gMLP

To the best of our knowledge, there no publications at the time of writing this paper that demonstrates the combination of 2D Convolution with gMLP. This research gap is presented in Figure 1. In May 2021, Google brain proposed gMLP, which performed comparable to Transformer based approach. The gMLP authors argue the need for self-Attention, and aim to propose an a smaller network without Attention. gMLP models spatial interactions. Inspired by progress in combing Convolution with Transformers to build models like ConViT and CViT, this works introduces Convolution with gMLP.

## 1. Methods & Results

### 2.1 Contributions

The contributions in this paper are

1. First to introduce Convolutions in gMLP. The new model is henceforth referred to as "Convolutional gated MLP" or CgMLP in this paper.
2. Implements Convolutional gated MLP
3. Experimental comparison of CgMLP vs gMLP on CIFAR dataset
4. Investigates how CgMLP learns by visualizing the feature maps
5. Contributes the source code of CgMLP with this paper

### 1.2 Convolution gated MLP

The neural network architecture of Convolution gated MLP is depicted in Fig 3. While the gMLP took in directly the 256x256 RGB image as 8x8 image patches, the CgMLP accepts visual feature maps. By adding 2D Convolution layers to the gMLP, the low level features are dynamically extracted and injected to gMLP blocks. The beauty of this approach is that a large 256x256 image could be condensed into a smaller feature map of 16x16 by CNN layers, thus enabling gMLP to process the entire image at one shot, rather than the current approach of dealing with image patch. This is particular useful to learn spatial interaction given gMLP doesn't use positional embedding. This opens door for making sense from features such as outline of the car as shown in Fig 6. The other benefit is since Convolution & MaxPool can absorb a set of neighborhood pixels of receptive 5x5 cell into a smaller feature, gMLP can work look at the entire image rather than in patches. The CgMLP simply consists of initial layers of 2D Convolution layers , and higher layers of gMLP layers. This is shown in Fig 5.

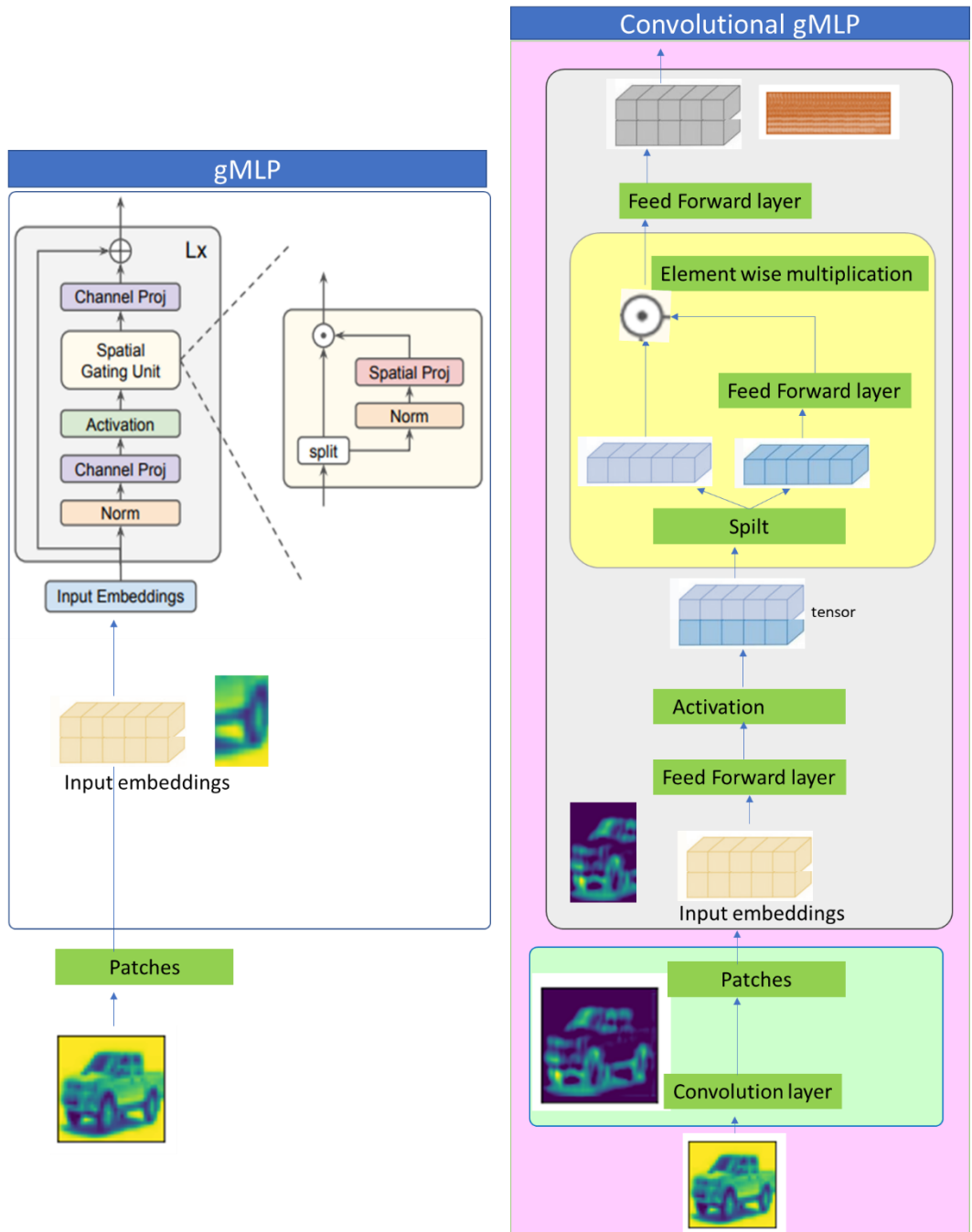

*Figure 3*

## 1.3 Comparison of gMLP vs CgMLP

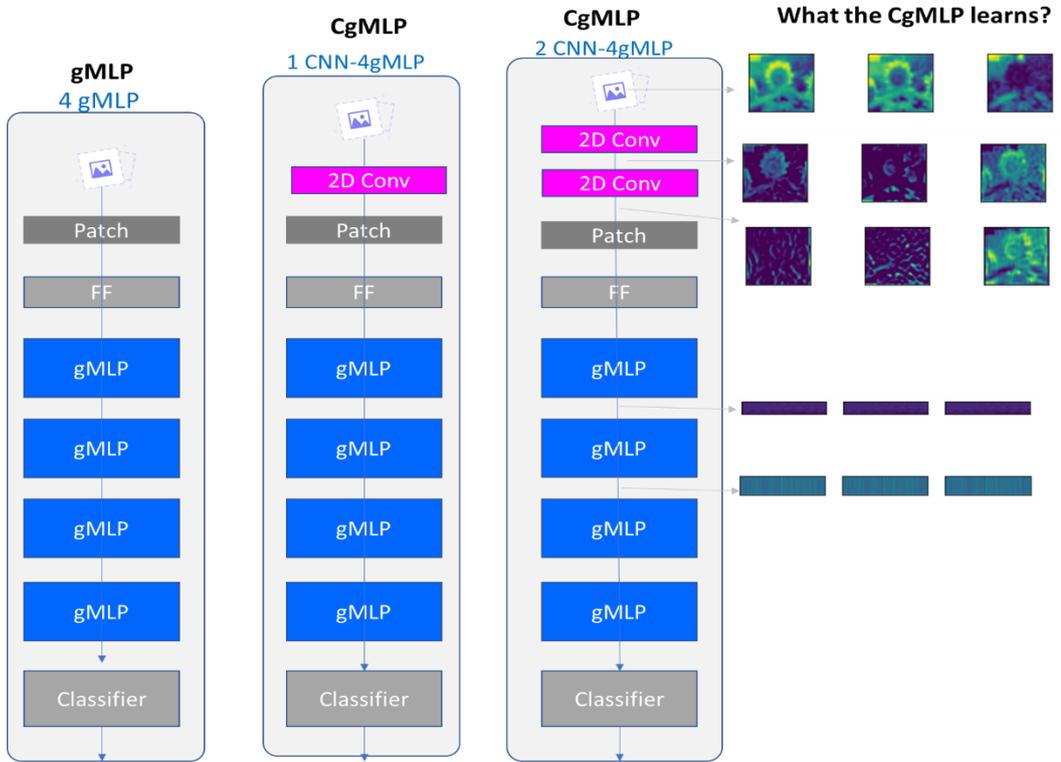

*Figure 4*

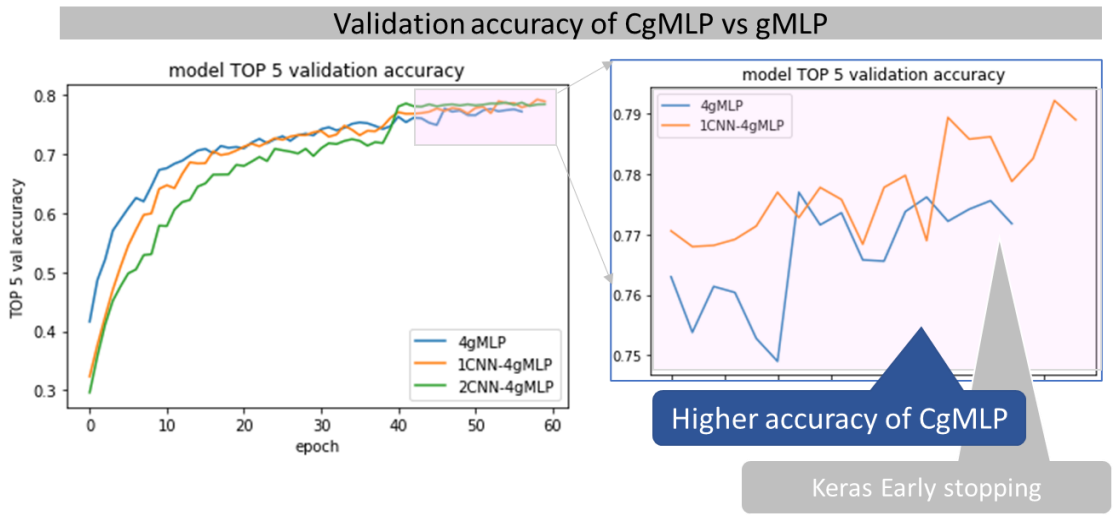

*Figure 5*

Experiments show CgMLP achieve equal or better validation accuracy than gMLP as shown in Fig 5. We trained 3 models on CIFAR-100. The 3 models are shown in Fig 4. The gMLP model had 4 blocks of gMLP units. Then we had 2 variations of CgMLP, one had 1 layer of CNN introduced, the other had 2 layers. As seen in Fig 5, the CgMLP achieves a competitive validation accuracy. Results show a right balance of CNN layers and gMLP would be beneficial. gMLP stopped training before 60 epochs due to TensorFlow early stopping set of validation accuracy, while CgMLP continues to train. This indicates the potential of CgMLP to generalize better!

### 1.4 How the Convolutional gated MLP learns?

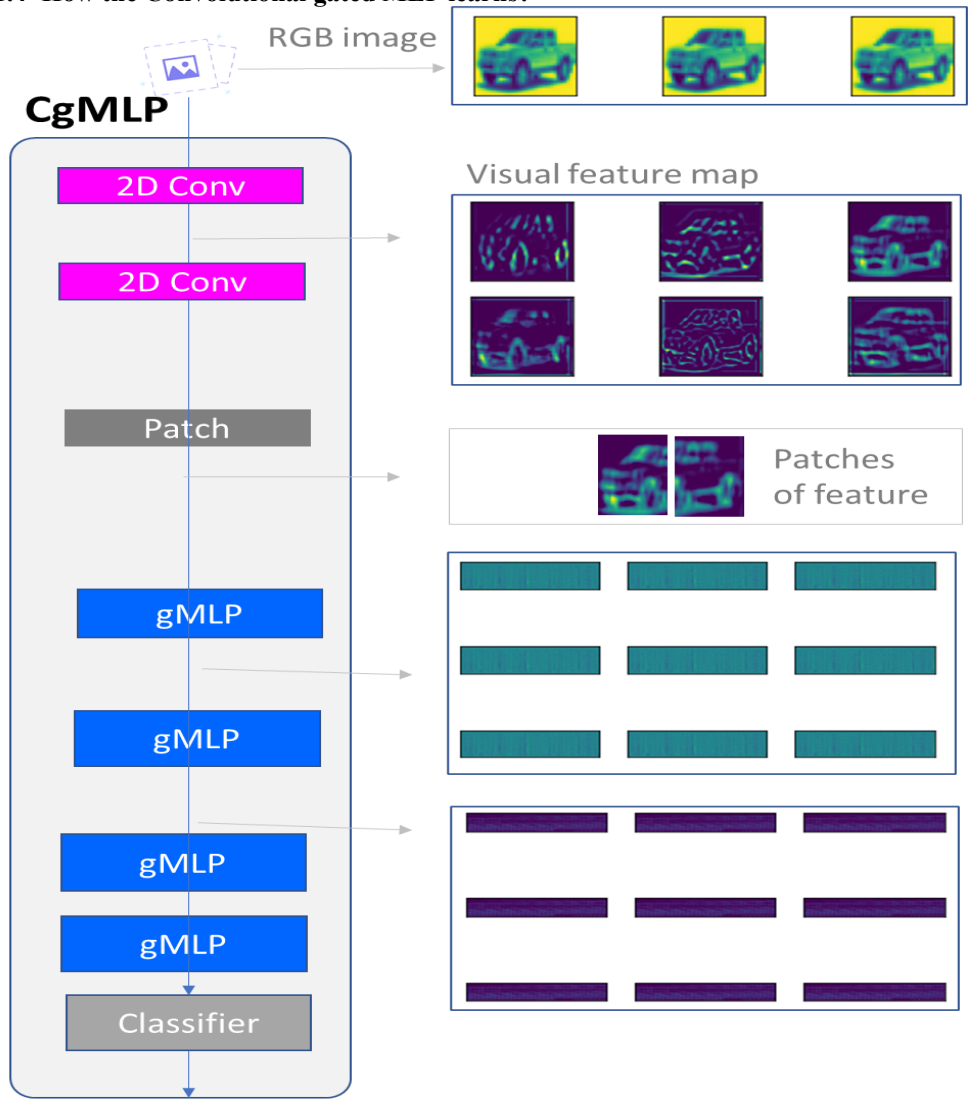

*Figure 6*

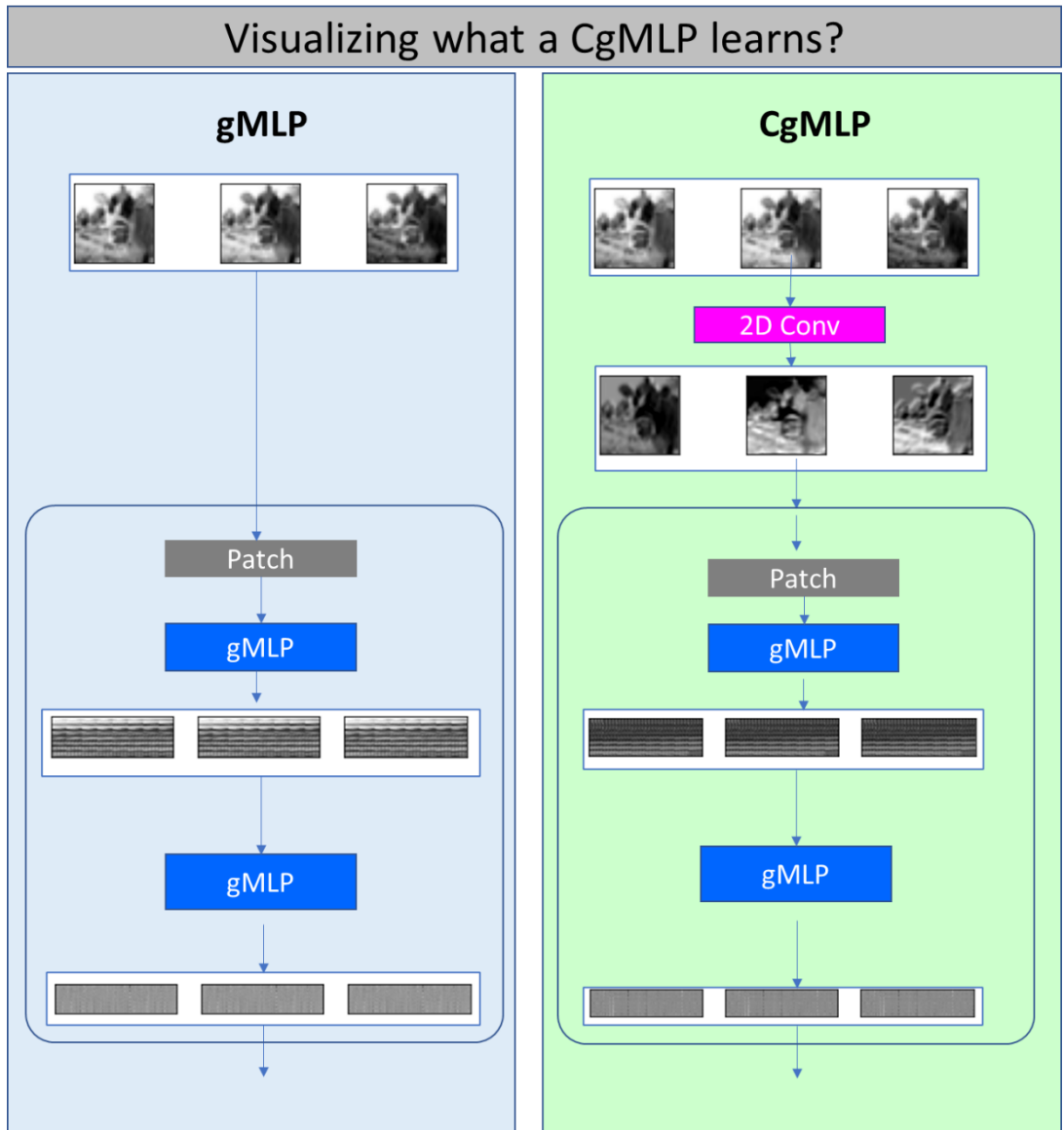

*Figure 7*

To gain insights of the proposed neural network architecture, we visualized the feature maps as the image is passed through various layers of model. It is interesting to note that the model learnt to extract certain set of visual features that are important for further downstream consumption. It is important to note that the Convolution layers are trained from scratch rather than borrow the learnings from IMGAGE pre-trained model. This meant the Convolution layer had the opportunity to focus on features that is most important for further downstream gMLP blocks. The CNN layers learnt to re-represent the input image to certain other representations.

There are few areas the gain the attention of the CgMLP network
1. Attention on Salient objects:
   - Key object features such as the focus on flower rather than the background garden. (Fig 4). The face of horse get the focus rather than the background playground (Fig 7).
2. Attention on outline of the object:
   - While any low level features could have been extracted, CgMLP extracted the outline of a car (Fig 8). In contrast to this, a VGG-16 learns multiple features such as color, textures, edges, etc. Due to selection of smaller number of filters on the CNN, the model just learnt to focus on the most important low level features.
3. Receptive field that lets the gMLP to focus on the whole rather than patches
   - The visual features from CNN can maxpolled to reduce the dimension of a 256x256 image to a 8x8 feature map, hence allowing the gMLP to process the entirety of the visual. This is important to improve the accuracy. In the gMLP, each patch is processed by series of gMLP layers. Then all the processed patches are finally pooled using a Pooling layer. In contrast, CgMLP can process the entire 256x256 receptive field at one go. Tunning the CgMLP to look at the entire image at one go can lead to improvements in accuracy.

## 1.5 gMLP vs CgMLP: spatial interactions vs feature channel interactions

The gating unit is gMLP has spatial projection layer. The gMLP authors defined spatial gating unit as $s(X) = X_1 * f\,W(X_2)$. The input image tensor is spilt into two tensors, $X_1$, $X_2$. The size of $X_1$ tensor is size 64x256 or (number of patches x embedding Dimension), when the number of image patches is 64. A spatial projection of trainable Weight, W of 1x256 matrix (when the embedding Dimension is set as 256).

In a CgMLP, this spatial projection can be flipped as channel projection across the channels of the feature map. This is shown in Fig 8. So if the CNN layer has 64 filters, there will be the 64 channels in the feature map. So the input feature map X can be split into two tensors $X_1$, $X_2$ in such a way so that there is modelling of channel interaction. So in CgMLP, the feature channel projection can be trainable Weight, W of 1x256 matrix. In short, CgMLP can learn channel interactions or spatial interactions based on the axis of splitting the tensors. This increase the fundamental power of CgMLP to model a either spatial interaction and channel interaction. So a network can consists of combination of two variants of CgMLP,
- CgMLP layer to model spatial interactions
- CgMLP layer to model channel interactions

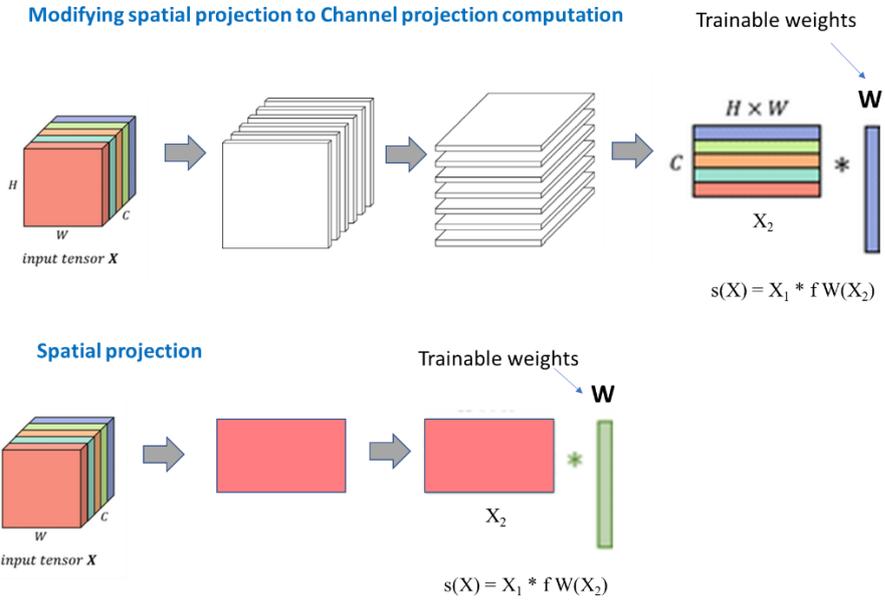

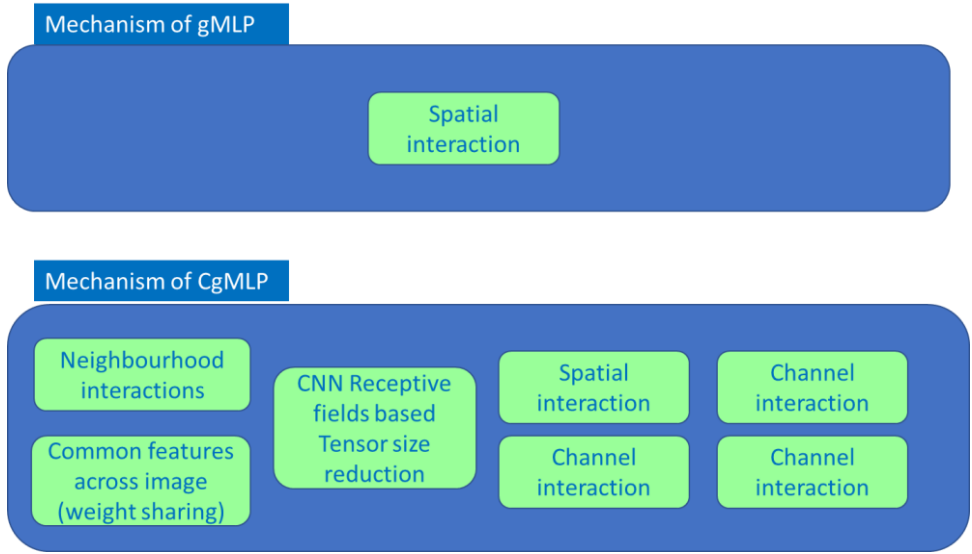

*Figure 8*

*Figure 9*

## 1.6 gMLP vs CgMLP: Inductive biases

|  | CgMLP | gMLP |
|---|---|---|
| 1. Model spatial interactions | Yes | Yes |
| 2. Model Channel interactions of feature map | Yes | No |
| 3. Learn to extract meaning from neighboring pixels | Yes | Partial |
| 4. Weight sharing to learn common features | Yes | No |
| 5. Model a smaller input embedding (by receptive field of CNN) | Yes | No |

*Table 2*

The opportunity to model more specialized interactions is possible with a CgMLP due to the factors listen in Table 2. A CgMLP neural network will employ all these 5 factors in combination to model the training data in a computer vision tasks. For example, a Convolution gMLP unit with spatial gating may be used in a layer, but other layers may include a Convolution gMLP with channel gating. Thus , CgMLP is a combination of different layers each specializing in a modelling a certain aspect of data. This vision of CgMLP architectural basis is shown in Fig 9.

## 1.7 Reproducible results & Source code availability

The source code of Convolution Gated MLP is contributed in open source at this URL, https://sites.google.com/view/convolutional-gated-mlp. The supplemental website also shows the visualization of how CgMLP works under the hood. The results are reproducible.

## 2. Summary

Since the introduction of gMLP by Google Brain in May 2021, not many publications have attended to combining gMLP with Convolution neural networks. This paper spotted this gap in knowledge, and proposed a novel architecture. The proposed architecture trained a network that combined 2D Convolution with gMLP as a sequence of neural network layers. This proposed Convolutional gated MLP (CgMLP) was implemented and trained on CIFAR-100. Experiments demonstrated the potential of blending Convolutions with gMLP. Experiments also tend to indicate that CgMLP could generalize, while the tendency of gMLP to over-fit on training data was reported by gMLP authors. Our experiment validated that CgMLP can train for more epochs than gMLP, as the experiment showed automatic early stopping on validation accuracy during Tensorflow training. While gMLP's gating unit performs spatial projection, a CgMLP's gating unit can be made to at projection across different channels of the CNN feature map. So gMLP mechanism is spatial interactions, CgMLP mechanism can feature channel interactions. The advantage of blending 2D Conv in a gMLP are many fold. The CgMLP architecture allowed the network to learn to attend to certain salient features, and freed up the gMLP blocks to look at the entire image through the eyes of receptive field of CNN. This allows for right sizing the neural network to open up the opportunity for tuning the network configuration for optimizing the generalization power of the model. By searching through the architecture search space, a design for ideal network configuration containing Convolution and gMLP can emerge. This opportunity to tune for generalization power is the key breakthrough this paper leads into for future direction.

# References

[1] S. d'Ascoli, H. Touvron, M. Leavitt, A. Morcos, G. Biroli, and L. Sagun, "ConViT: Improving Vision Transformers with Soft Convolutional Inductive Biases," *arXiv:2103.10697 [cs, stat]*, Jun. 2021, Accessed: Oct. 30, 2021. [Online]. Available: https://arxiv.org/abs/2103.10697.
[2] H. Liu, Z. Dai, D. R. So, and Q. V. Le, "Pay Attention to MLPs," *arXiv:2105.08050 [cs]*, Jun. 2021, Accessed: Oct. 30, 2021. [Online]. Available: https://arxiv.org/abs/2105.08050.
[3] H. Wu *et al.*, "CvT: Introducing Convolutions to Vision Transformers," *arXiv:2103.15808 [cs]*, Mar. 2021, Accessed: Oct. 17, 2021. [Online]. Available: https://arxiv.org/abs/2103.15808.
[4] C. Edwards, "The best of NLP," *Communications of the ACM*, vol. 64, no. 4, pp. 9–11, Apr. 2021, doi: 10.1145/3449049.
[5] X. Li *et al.*, "Oscar: Object-Semantics Aligned Pre-training for Vision-Language Tasks," *Computer Vision – ECCV 2020*, pp. 121–137, 2020, doi: 10.1007/978-3-030-58577-8_8.
[6] Z. Wang, J. Yu, A. W. Yu, Z. Dai, Y. Tsvetkov, and Y. Cao, "SimVLM: Simple Visual Language Model Pretraining with Weak Supervision," *arXiv:2108.10904 [cs]*, Aug. 2021, Accessed: Oct. 30, 2021. [Online]. Available: https://arxiv.org/abs/2108.10904.
[7] H. Le, D. Sahoo, N. F. Chen, and S. C. H. Hoi, "Multimodal Transformer Networks for End-to-End Video-Grounded Dialogue Systems," *Proceedings of the 57th Annual Meeting of the Association for Computational Linguistics*, pp. 5612–5623, 2019, doi: 10.18653/v1/P19-1564.
[8] Y. Xu, Q. Zhang, J. Zhang, and D. Tao, "ViTAE: Vision Transformer Advanced by Exploring Intrinsic Inductive Bias," *arXiv:2106.03348 [cs]*, Jul. 2021, Accessed: Oct. 30, 2021. [Online]. Available: https://arxiv.org/abs/2106.03348.
[9] A. Vaswani *et al.*, "Attention Is All You Need." [Online]. Available: https://proceedings.neurips.cc/paper/2017/file/3f5ee243547dee91fbd053c1c4a845aa-Paper.pdf.
[10] L. Wu, Z. Zhu, and W. E, "Towards Understanding Generalization of Deep Learning: Perspective of Loss Landscapes," *arXiv:1706.10239 [cs, stat]*, Nov. 2017, Accessed: Oct. 30, 2021. [Online]. Available: https://arxiv.org/abs/1706.10239.
[11] A. Dosovitskiy *et al.*, "An Image is Worth 16x16 Words: Transformers for Image Recognition at Scale," *arXiv:2010.11929 [cs]*, Oct. 2020, Accessed: Oct. 30, 2021. [Online]. Available: https://arxiv.org/abs/2010.11929v1.

+rajagopal.motivate@gmail.com Indian Institute of Technology, Madras
++ gvan.nirmala@gmail.com (correspondence) Queen Mary's College